\documentclass[sigconf]{acmart}
\AtBeginDocument{%
  }

\copyrightyear{2025}
\acmYear{2025}
\setcopyright{cc}
\setcctype{by}
\acmConference[CIKM '25]{Proceedings of the 34th ACM International Conference on Information and Knowledge Management}{November 10--14, 2025}{Seoul, Republic of Korea}
\acmBooktitle{Proceedings of the 34th ACM International Conference on Information and Knowledge Management (CIKM '25), November 10--14, 2025, Seoul, Republic of Korea}\acmDOI{10.1145/3746252.3761318}
\acmISBN{979-8-4007-2040-6/2025/11}

\usepackage{xcolor} 
\usepackage{enumitem}
\usepackage{algpseudocode}
\usepackage{multirow}
\usepackage{tabularx}
\usepackage{soul}
\usepackage{url}
\usepackage{amsthm}
\usepackage{algorithm}
\usepackage{algorithmicx}
\usepackage{dsfont}
\begin{document}

%%
%% The "title" command has an optional parameter,
%% allowing the author to define a "short title" to be used in page headers.
\title{LatentExplainer: Explaining Latent Representations in Deep Generative Models with Multimodal Large Language Models}

%%
%% The "author" command and its associated commands are used to define
%% the authors and their affiliations.
%% Of note is the shared affiliation of the first two authors, and the
%% "authornote" and "authornotemark" commands
%% used to denote shared contribution to the research.
\author{Mengdan Zhu}
\affiliation{%
  \institution{Emory University}
  \department{Department of Computer Science}
  \city{Atlanta}
  \state{GA}
  \country{USA}
}
\email{mengdan.zhu@emory.edu}

\author{Raasikh Kanjiani}
\affiliation{%
  \institution{Emory University}
  \department{Department of Computer Science}
    \city{Atlanta}
  \state{GA}
  \country{USA}}
\email{raasikh.kanjiani@emory.edu}

\author{Jiahui Lu}
\affiliation{%
  \institution{University College London}
  \department{Department of Computer Science}
    \city{London}
  \country{United Kingdom}}
\email{jiahui.lu.24@ucl.ac.uk}

\author{Andrew Choi}
\affiliation{%
  \institution{Emory University}
  \department{Department of Computer Science}
    \city{Atlanta}
  \state{GA}
  \country{USA}}
\email{andrew.jaemin.choi@emory.edu}

\author{Qirui Ye}
\affiliation{%
  \institution{Emory University}
  \department{Department of Computer Science}
    \city{Atlanta}
  \state{GA}
  \country{USA}}
\email{qirui.ye@emory.edu}

\author{Liang Zhao}
\authornote{Corresponding author.}
\affiliation{%
  \institution{Emory University}
  \department{Department of Computer Science}
    \city{Atlanta}
  \state{GA}
  \country{USA}}
\email{liang.zhao@emory.edu}

%%
%% By default, the full list of authors will be used in the page
%% headers. Often, this list is too long, and will overlap
%% other information printed in the page headers. This command allows
%% the author to define a more concise list
%% of authors' names for this purpose.
\renewcommand{\shortauthors}{Mengdan Zhu et al.}
\renewcommand{\shorttitle}{LatentExplainer}
%%
%% The abstract is a short summary of the work to be presented in the
%% article.
\begin{abstract}
Deep generative models like VAEs and diffusion models have advanced various generation tasks by leveraging latent variables to learn data distributions and generate high-quality samples. Despite the field of explainable AI making strides in interpreting machine learning models, understanding latent variables in generative models remains challenging. This paper introduces \textit{LatentExplainer}\footnote{The code is available at ~\url{https://github.com/mengdanzhu/LatentExplainer}.}, a framework for automatically generating semantically meaningful explanations of latent variables in deep generative models. \textit{LatentExplainer} tackles three main challenges: inferring the meaning of latent variables, aligning explanations with inductive biases, and handling varying degrees of explainability. Our approach perturbs latent variables, interprets changes in generated data, and uses multimodal large language models (MLLMs) to produce human-understandable explanations. We evaluate our proposed method on several real-world and synthetic datasets, and the results demonstrate superior performance in generating high-quality explanations for latent variables. The results highlight the effectiveness of incorporating inductive biases and uncertainty quantification, significantly enhancing model interpretability.
\end{abstract}

%%
%% The code below is generated by the tool at http://dl.acm.org/ccs.cfm.
%% Please copy and paste the code instead of the example below.
%%

%%
%% Keywords. The author(s) should pick words that accurately describe
%% the work being presented. Separate the keywords with commas.
%% A "teaser" image appears between the author and affiliation
%% information and the body of the document, and typically spans the
%% page.

%%
%% This command processes the author and affiliation and title
%% information and builds the first part of the formatted document.
\maketitle

\section{Introduction}
\begin{figure}[t]
\begin{center}
\includegraphics[width=0.48\textwidth]{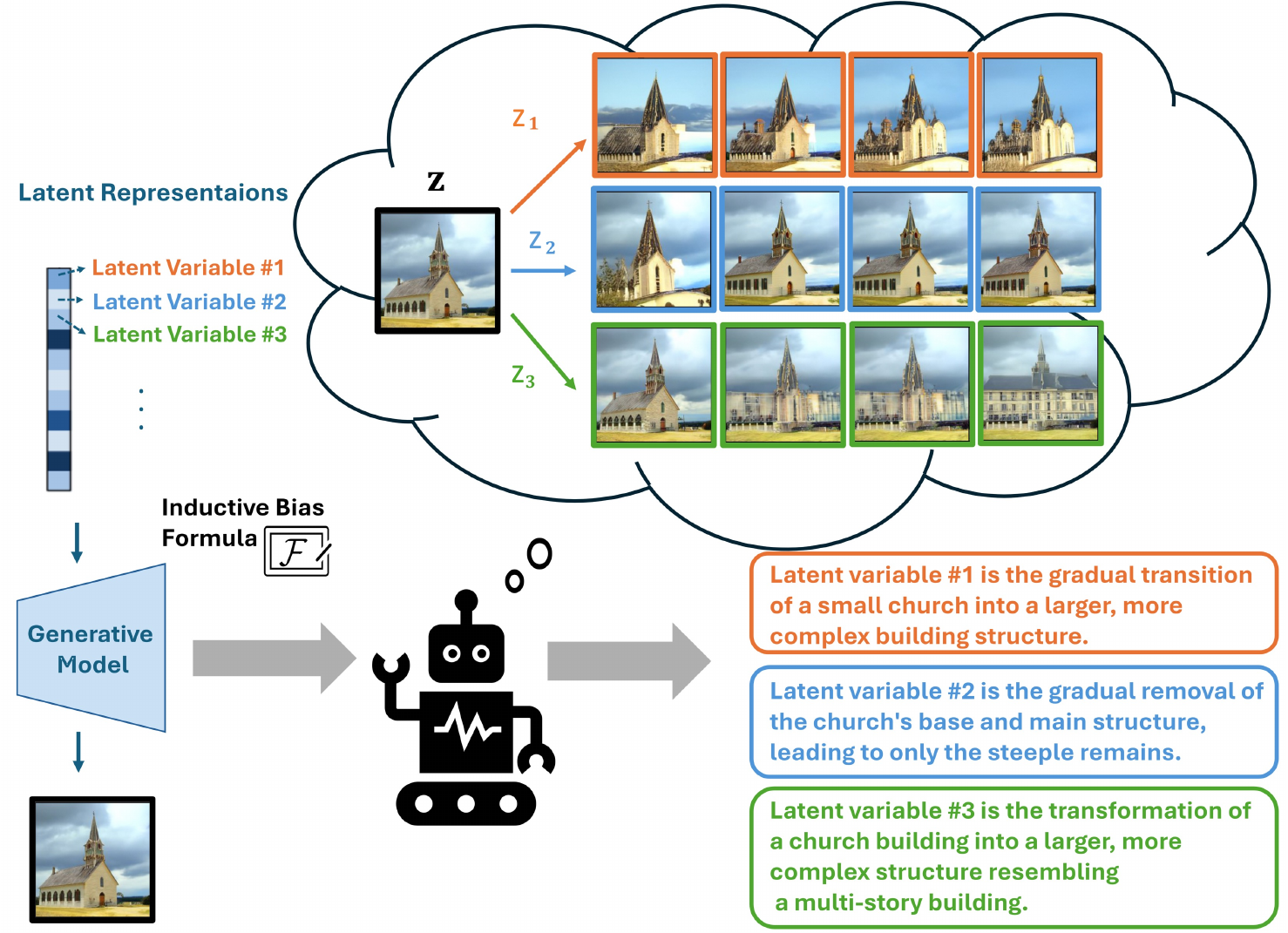}
\caption{Illustration of how a pretrained generative model, guided by inductive bias formulas, automatically decodes and interprets meaningful latent dimension semantics.}

\label{fig:introduction_figure}
\end{center}
\end{figure}

Deep generative models, such as Variational Autoencoders (VAEs)~\cite{kingma2013auto} and diffusion models~\cite{rombach2022high}, have become a state-of-the-art approach in various generation tasks~\cite{ho2022cascaded,yang2023diffusion}. These methods effectively leverage \emph{latent variables} to learn underlying data distributions and generate high-quality samples by capturing the underlying structure of high-dimensional data in a low-dimensional semantic space. As the latent variables represent all the information in a lower dimension, they can be considered as an effective abstraction of key factors in the data. Therefore, it is critical to develop methods for automatically decomposing and explaining meaningful latent dimension semantics given a pretrained generative model and its inherent inductive biases, as illustrated in Figure~\ref{fig:introduction_figure}. \emph{Inductive biases} are often enforced over latent variables in deep generative models. For instance, disentanglement is a rule of thumb which enforces orthogonality among different latent variables~\cite{ding2020guided}. Moreover, sometimes latent variables can be grouped, leading to combination bias~\cite{klys2018learning}. More recently, the desire for controllability in deep generative models, where latent variables are associated with specific properties of interest~\cite{wang2024controllable}, has given rise to conditional bias. Incorporating inductive biases aligned with the actual facts can reduce the hallucination in explaining the latent variables in deep generative models \cite{wu2024no}.
The field of explainable artificial intelligence (XAI) has extensively investigated the interpretation of machine learning models~\cite{adadi2018peeking,zhu2021development}. However, interpreting latent variables in deep generative models remains underexplored. Machine learning model explanation, a.k.a., \emph{post-hoc explanation}, can be categorized into global and local explanations~\cite{gao2024going}. Global explanations focus on elucidating the entire model, while local explanations target the reasoning behind specific predictions. Global explanations are more challenging, with existing work mostly emphasizing attributions to identify which features are most important for model decision-making~\cite{saleem2022explaining}. The missing piece is understanding the meaning of features when they are unknown, which is very common in deep generative models. 
More recently, one category of global explanation methods, called \emph{concept-based} explanations, aims to generate more human-understandable concepts as explanations~\cite{poeta2023concept}. However, current concept-based methods often rely on human heuristics or predefined concept and feature space, limiting the expressiveness of the explanations and falling short of achieving truly automatic explanation generation~\cite{koh2020concept,bai2022concept}.

Despite the progress in XAI, interpreting latent variables in deep generative models presents significant challenges. \textbf{First}, these variables are not grounded in real-world concepts, and the black-box nature of the models prevents us from inferring the meaning of latent variables from observations. \textbf{Second}, explanations must adhere to the inductive biases imposed on the latent variables, which is essential yet difficult to ensure. For example, in disentangled latent variables, the semantic meanings should be orthogonal. \textbf{Third}, different latent variables have varying degrees of explainability. Some may be trivial to data generation and intrinsically lack semantic meaning. It is crucial to identify which latent variables are explainable and which do not need explanations. 

To address the aforementioned challenges, we propose \textit{LatentExplainer}, a novel and generic framework that automatically generates semantically-meaningful explanations of latent variables in deep generative models. 
Specifically, to explain these variables and work around the black-box nature of the models, we propose to perturb each latent variable and explain the resulting changes in the generated data. Specifically, we perturb and decode each manipulated latent variable to produce the corresponding sequence of generated data samples. The trend in the sequence is leveraged to reflect the semantics of the latent variable to be explained. To align explanations with the intrinsic nature of the deep generative models, we design a generic framework that formulates inductive biases on the Bayesian network of latent variable models into textual prompts. These prompts are understandable to large foundation models and humans. To handle the varying degree of explainability in latent variables, we propose to measure the confidence of the explanations by estimating their uncertainty. This approach assesses whether the latent variables are interpretable and selects the most consistent explanations, ensuring accurate and meaningful interpretation of the latent variables.

\section{Related Work}
\label{sec:related work}

\noindent\textbf{Deep Generative Models.}
Deep generative models are essential for modeling complex data distributions. Variational Autoencoders (VAEs) are prominent in this area, introduced by Kingma and Welling \cite{kingma2013auto}. VAEs encode input data into a latent space and decode it back, optimizing a balance between reconstruction error and the Kullback-Leibler divergence \cite{rezende2014stochastic}. They have diverse applications, including image generation \cite{yan2016attribute2image}, and anomaly detection \cite{an2015variational}.

Diffusion models, proposed by Ho et al. \cite{ho2020denoising}, generate data by a diffusion process that gradually adds noise to the data and then learns to reverse this process to recover the original data. These models have achieved high-fidelity image generation, surpassing generative adversarial networks (GANs) in quality and diversity. Latent diffusion models allow the model to operate in a lower-dimensional space, which significantly reduces computational requirements while maintaining the quality of the generated samples \cite{rombach2022high}. Advances have made them applicable to text-to-image synthesis \cite{nichol2021glide}, and audio generation \cite{kong2020diffwave}.

\noindent\textbf{Latent Variable Manipulation and Explanations.}
Manipulating latent variables in generative models like VAEs and diffusion models is an important technique for editing and enhancing generated images. A key method is latent traverse, which involves traversing different values of latent variables to achieve diverse manipulations in the generated outputs. This technique allows for precise control over the attributes in generated images, enabling adjustments~\cite{chen2016infogan}. For example, latent traverse has been effectively employed to disentangle and control various attributes in generated images~\cite{brock2016neural,zhu2016generative}. However, latent traverse is often used for visualization and editing purposes. It has not yet been widely explored as a tool for explaining the underlying latent space. Their explanations primarily rely on using predefined training attributes as text labels or manually adding explanations~\cite{shen2020interpreting,esser2020disentangling}. Some concept-based models control latent variables in generative models using category concepts to generate data~\cite{tran2022unsupervised,bouchacourt2018multi}. However, these approaches cannot automatically generate free-form textual explanations.

% Other methods for explaining latent variables include feature importance techniques, where the significance of latent dimensions is quantified to understand their impact on generation quality \cite{ridgeway2018learning}.

More recently, Multimodal Large Language Models (MLLMs) integrate diverse data modalities, enhancing their ability to understand and generate complex information~\cite{yin2023survey,bai2024beyond}. Notable models include GPT-4o, which extends GPT-4v with better visual capabilities \cite{openai2024gpt4o}, and Gemini that is a family of highly capable multimodal models \cite{team2024gemini}. We envision leveraging MLLMs to automatically generate explanations for latent variables and incorporate their inductive biases to reduce hallucination. Our work focuses on: (1) how to decompose the inductive bias formulas to automate the manipulation of latent variables, (2) how to develop prompts that are aligned with the underlying inductive biases of generative models and can be easily understood by MLLMs, and (3) how to evaluate the quality of the generated explanations for latent variables.

\begin{figure*}[t]
\begin{center}
\includegraphics[width=\textwidth]{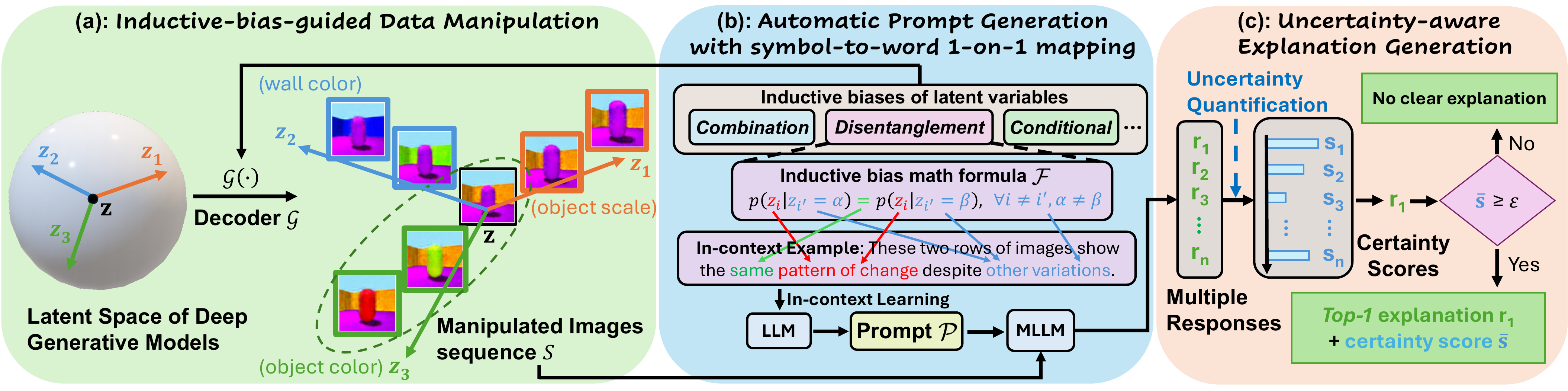}
\caption{The overview of our proposed framework \textit{LatentExplainer} (a) \textit{Inductive-bias-guided Data Manipulation} generates image sequences by manipulating latent variables with predefined biases; (b) \textit{Automatic Prompt Generation with symbol-to-word mapping} uses these images and formulas to create prompts for an MLLM to produce explanations; (c) \textit{Uncertainty-aware Explanation Generation} evaluates multiple responses from the MLLM, selecting the most consistent explanation with a certainty score.}
\label{fig:overview}
\end{center}
\end{figure*}

\section{Preliminaries and Problem Formulation}
\label{sec:problem_formulation}

\noindent\textbf{Deep Generative Models.}
These are a class of models that learn a mapping between observations $\mathbf{x}$ and underlying latent factors $\mathbf{z}$. These models are widely used in deep generative models such as VAEs and latent diffusion models. VAEs, for instance, introduce a probabilistic approach to encoding data by maximizing the evidence lower bound (ELBO)~\cite{kingma2013auto}:
\begin{displaymath}
\mathcal{L}_{\text{ELBO}} = \mathbb{E}_{q(\mathbf{z}|\mathbf{x})}[\log p(\mathbf{x}|\mathbf{z})] - \text{KL}(q(\mathbf{z}|\mathbf{x}) \| p(\mathbf{z})),   
\label{eq:elbo_vae}
\end{displaymath}
where KL stands for Kullback–Leibler divergence. 

Latent diffusion models further refine the generation process by iteratively refining noise into structured data~\cite{rombach2022high}. These models effectively capture the underlying structure of data in a low-dimensional semantic space. Latent variable manipulation of diffusion models aims at transversing the latent variables $\mathbf{z}$ along the semantic latent direction $\mathbf{z_i}$. The perturbed vector $\tilde{\mathbf{z}}=\mathbf{z}+\gamma\left[\mathcal{G}(\mathbf{z}+\mathbf{z_i})-\mathcal{G}(\mathbf{z})\right]$, where $\gamma$ is a hyper-parameter controlling the strength and $\mathcal{G}$ is a diffusion decoder \cite{park2023understanding}. An image sequence can then be generated by $\mathcal{G}(\mathbf{\tilde{z}})$. The perturbations in the semantic latent direction lead to semantic changes in the generated image sequence.   

\noindent\textbf{Inductive Bias in Latent Variables.}
Inductive biases are usually imposed on latent variables to enhance the performance and interpretability of deep generative models. These inductive biases in deep generative models can be categorized into three common types:
\emph{Disentanglement Bias:} It enforces orthogonality among different latent variables, ensuring that each latent variable captures a distinct factor of variation in the data~\cite{ding2020guided}.
\emph{Combination bias:} Sometimes latent variables are grouped, leading to biases in how they interact and combine to represent complex data structures~\cite{klys2018learning}.
\emph{Conditional Bias:} It emphasizes the relationship between specific properties of interest and the corresponding latent variables \cite{wang2024controllable}. 

% These inductive biases are crucial for making the latent representations more interpretable and controllable, thus enhancing the overall utility of deep generative models.
\noindent\textbf{Problem Formulation.} We assume a dataset $\mathcal{D}$, where each sample consists of $x$ or $(x,y)$, with $x \in \mathbb{R}^N$ and $y=\left\{y_k \in\right.$ $\mathbb{R}\}_{k=1}^K$ as $K$ properties of $x$. The dataset $\mathcal{D}$ is generated by M latent variables $z_i$, where $i \in \{1,\ldots, M\}$. $z_i$ can be an individual latent dimension in $z$ in the VAE models or a latent vector in the diffusion models.  Suppose we are given a generative model with a set of formulas $\mathcal{F}$ with respect to $z_i$, where $\mathcal{F}$ represents an inductive bias that the generative model must satisfy. Our goal is to derive a textual sequence that explains the semantic meanings of $z_i$.

\section{Proposed Method}
\subsection{Overview of \textit{LatentExplainer}}
This paper focuses on the tasks in explaining the semantics of latent variables $\{{z}_i\}_{i=1}^M$ in deep generative models. To interpret the semantics of latent variables and work around the blackbox nature of deep generative models, we propose to perturb each $z_i$ and explain the change it imposes on the generated data. To solve these challenges, we propose a novel \textit{LatentExplainer} scheme. The pseudo-code of this whole scheme can be found in Algorithm~\ref{alg:latentexplainer}. When explaining latent variable models, it is crucial to fully leverage and align with the prior knowledge about them. To do this, we design a generic framework that can automatically formulate inductive bias of generative models into textual prompts. Specifically, we have summarized three common inductive biases and designed their symbol-to-word one-on-one prompts $\mathcal{P}$ (Section \ref{sec:bias}). Our scheme can adaptively convert a user-provided inductive bias formulas $\mathcal{F}$ into a corresponding prompt $\mathcal{P}$ to provide more accurate explanations of the latent variables in Figure~\ref{fig:overview}(b) (Section \ref{sec:prompt}). We decompose the inductive bias to guide the perturbation of $z_{i}$ and subsequently decode the manipulated latent variables into generated data that are perceptible by humans such as images. Through a series of perturbations on $z_i$, a sequence of generated data samples can be obtained to reflect the changes in $z_i$ in Figure~\ref{fig:overview}(a) (Section \ref{sec:manipulate}). Eventually, the explanations are selected through an uncertainty quantification approach to assess whether the latent variables are interpretable and select the most consistent explanations in Figure~\ref{fig:overview}(c) (Section \ref{sec:uncertainty}). 

\subsection{Inductive-bias-guided Prompt Framework}
\label{sec:bias}

\subsubsection{Generic Framework}

In this section, we propose a generic framework that can verbalize the inductive bias in deep generative models into prompts for better latent variable explanations. The prevalent inductive biases in deep generative models are categorized into three types: disentanglement bias, conditional bias, and combination bias. 

Our framework proposes a principled, automatic way that translate the mathematical expression to textual prompts. The prompts include adaptive prompts and a fixed ending. The adaptive prompts are converted from the inductive bias formulas. The formulas contain mathematical symbols that consist of mathematical variables and mathematical operators. We use the same color to represent the correspondence between mathematical symbols in the formulas and the text in the prompts. The translation mechanism of adaptive prompts is shown in Table~\ref{table:lookup}. 
\begin{table}[t]
\caption{Lookup table for symbol-to-word mapping.}
\begin{tabular}{|c|c|p{3cm}|}
\hline
Grammar \# & Symbol & Prompt \\ \hline
1          &   $p$(\textcolor{purple}{$z_i$} $\mid \cdot$)     &   \textcolor{purple}{pattern of change}     \\ \hline
2          &  $p(z_i \mid$\textcolor{brown}{$z_{i^{\prime}}$}), \textcolor{brown}{$\forall i \neq i^{\prime}$}       & \textcolor{brown}{other variations}       \\ \hline
3          &   \textcolor{orange}{$p_k$}       &   \textcolor{orange}{property of interests}      \\ \hline
4          &   \textcolor{magenta}{$G$}      &    \textcolor{magenta}{a group}     \\ \hline
5          &   \textcolor{blue}{$\in$}      &    \textcolor{blue}{associated with}    \\ \hline
6          & \textcolor{blue}{$\notin$}       &   \textcolor{blue}{not associated with}     \\ \hline
7          &  \textcolor{teal}{=}       & \textcolor{teal}{same}       \\ \hline
8          &  \textcolor{cyan}{$\neq$}      &  \textcolor{cyan}{change}      \\ \hline
\end{tabular}
\label{table:lookup}
\end{table}
Fixed Ending: What is the pattern of change? Write in a sentence. If there is no clear pattern, just write "No clear explanation".

\subsubsection{From Disentanglement Bias to Prompts}
\label{sssec:disentanglement_bias}

Disentanglement bias refers to the model's ability to separate independent factors in the data~\cite{ding2020guided,wu2023uncovering}. The formula representing this bias focuses on ensuring that different latent variables correspond to different independent underlying factors. Independent factors would be invariant with respect to one another \cite{ridgeway2016survey}. By disentangling these factors, researchers can better understand the underlying structure of the data and improve the model's performance on tasks such as representation learning.

\textbf{Formula:} 
\begin{equation*}
    p(\textcolor{purple}{z_i} \mid \textcolor{brown}{z_{i^{\prime}} =\alpha}) \textcolor{teal}{=} p(\textcolor{purple}{z_i} \mid \textcolor{brown}{z_{i^{\prime}} =\beta}), \textcolor{brown}{\forall i \neq i^{\prime}, \alpha \neq \beta}.
\end{equation*}
The above formula is translated into the following prompting using the grammar \#1,2,7.

\textbf{Prompt:} These two rows of images show the \textcolor{teal}{same} \textcolor{purple}{pattern of change} despite \textcolor{brown}{other variations}.

\subsubsection{From Combination Bias to Prompts}
\label{sssec:combination_bias}

Combination bias involves understanding how different latent variables interact within groups and remain independent across groups~\cite{klys2018learning}. This bias is significant as it helps in identifying how combinations of factors contribute to the overall data generation process. Recognizing these interactions enables researchers to design models that can generate more complex and realistic data by capturing intricate relationships within the data.

% Denote $z_{i}$ and $z_{{i^{\prime}}}$ as two latent variables in group $G$, and $z_{j}$ as the latent variable in $G^{\prime}$.
\begin{itemize}[leftmargin=*]

\item No inter-group correlation: 

\textbf{Formula:}
\begin{equation*}
    \resizebox{\linewidth}{!}{$p\left (\textcolor{purple}{z_{i}} \mid \textcolor{brown}{z_{j}=\alpha}\right)\textcolor{teal}{=}p\left(\textcolor{purple}{z_{i }} \mid \textcolor{brown}{z_{j}=\beta}\right), \forall \textcolor{purple}{z_i} \textcolor{blue}{\in} \textcolor{magenta}{G}, \textcolor{brown}{z_j} \textcolor{blue}{\in} \textcolor{magenta}{G^{\prime}}, \textcolor{brown}{G \neq G^{\prime}}, \textcolor{brown}{\alpha \neq \beta}.$} 
\end{equation*}
\end{itemize}

The above formula is translated into the following prompting using the grammar \#1,2,4,5,7.

\textbf{Prompt:} \textcolor{purple}{The pattern of change} is \textcolor{blue}{associated with} \textcolor{magenta}{a group}. The first two rows of images show the \textcolor{teal}{same} \textcolor{purple}{pattern of change} despite \textcolor{brown}{other variations} in \textcolor{magenta}{another group}.
\begin{itemize}[leftmargin=*]
\item Intra-group correlation:

\textbf{Formula:}
\begin{equation*}
\begin{split}
 & p\left(\textcolor{purple}{z_{i }} \mid \textcolor{brown}{z_{i^{\prime}}=\alpha}\right) \textcolor{cyan}{\neq} p\left(\textcolor{purple}{z_{i}} \mid \textcolor{brown}{z_{i^{\prime}}=\beta}\right),  \forall \textcolor{purple}{z_i}, \textcolor{brown}{z_i^{\prime}} \textcolor{blue}{\in} \textcolor{magenta}{G}, \textcolor{brown}{i \neq i^{\prime}, \alpha \neq \beta}.
\end{split}
\end{equation*}
\end{itemize}
The above formula is translated into the following prompting using the grammar \#1,2,4,5,8.

\textbf{Prompt:} \textcolor{purple}{The pattern of change} is \textcolor{blue}{associated with} \textcolor{magenta}{a group}. \textcolor{purple}{The pattern of change} in the last two rows of images should \textcolor{cyan}{change} given \textcolor{brown}{other variations}.

% \begin{figure*}[t]
% \begin{center}
% \includegraphics[width=\textwidth]{Figures/prompt.pdf}
% \caption{3-shot demonstrations for prompting LLMs to generate $\mathcal{P}$ based on $\mathcal{F}$.}
% \label{fig:prompt}
% \end{center}
% \end{figure*}

\subsubsection{From Conditional Bias to Prompts}
\label{sssec:conditional_bias}

Conditional bias focuses on the relationship between specific properties of interest and the corresponding latent variables~\cite{wang2024controllable}. This bias is important because it allows models to generate data conditioned on particular attributes, enhancing the model's ability to produce targeted and controlled outputs.

\textbf{Formula:}
% Let $p_k$ be the property of interest. The latent variable $z_i$ in conditional generative models tends to align with $p_k$.
\begin{equation*}
 p(\textcolor{purple}{z_i} \mid \textcolor{brown}{p_k =\alpha})\textcolor{cyan}{\neq}p(\textcolor{purple}{z_i} \mid \textcolor{brown}{p_k =\beta}), \forall \textcolor{purple}{z_i} \textcolor{blue}{\in} \textcolor{magenta}{G}\textcolor{orange}{_{k}}, \textcolor{brown}{\alpha\ne\beta}.
\end{equation*}
The above formula is translated into the following prompting using the grammar \#1,2,3,4,5,8.

\textbf{Prompt:} If \textcolor{purple}{the pattern of change} is \textcolor{blue}{associated with} the \textcolor{magenta}{group} of \textcolor{orange}{the property of interest}, this image sequence will \textcolor{cyan}{change} as \textcolor{brown}{other variations in $[\text{property}_k]$}.

There may exists a latent variable $z_j$ that are independent of $p_k$: 

\textbf{Formula:}
\begin{equation*}
p(\textcolor{purple}{z_j} \mid \textcolor{brown}{p_k =\alpha})\textcolor{teal}{=}p(\textcolor{purple}{z_j} \mid \textcolor{brown}{p_k =\beta}), \forall \textcolor{purple}{z_j} \textcolor{blue}{\notin} \textcolor{magenta}{G}\textcolor{orange}{_{k}}, \textcolor{brown}{\alpha \neq \beta}.
\end{equation*}
The above formula is translated into the following prompting using the grammar \#1,2,3,4,6,7.

\textbf{Prompt:} If \textcolor{purple}{the pattern of change} is \textcolor{blue}{not associated with} the \textcolor{magenta}{group} of \textcolor{orange}{the property of interest}, this image sequence will \textcolor{teal}{remain constant} despite \textcolor{brown}{other variations in $[\text{property}_k]$}.

\begin{algorithm}[t]
\caption{The \textit{LatentExplainer} Algorithm}
\label{alg:latentexplainer}
\begin{algorithmic}[1]
\State \textbf{Input:} Inductive Bias Formula(s) $\mathcal{F}$,  Optional Information About the Symbol $\mathcal{I}$
\State \textbf{Require:} Few-Shot Examples $\mathcal{H}$, Pre-trained LLM $\pi_\theta$, Pre-trained MLLM $\pi_{\theta^\prime}$, Generative Model Decoder $\mathcal{G}$

\State \textbf{Output:} Final Explanation $\hat{r}$
    \State symbols $\gets$ \Call{ExtractSymbols}{$\mathcal{F}$} 
    \State semantics $\gets \emptyset$   
    \For{symbol in symbols}  
        \State semantic $\gets$ $\pi_{\theta}({\text{symbol},\mathcal{H},\mathcal{I}})$
        \State semantics[symbol] $\gets$ semantic
    \EndFor
    \State $\mathcal{P} = \pi_{\theta}(\mathcal{F}, \mathcal{H}, \text{semantics})$  \Comment{Generate inductive bias prompt}

\State $\tilde{z} \gets \textsc{ManipulateLatent}(z, \mathcal{F}, \mathcal{P})$ \Comment{Latent variable perturbation}
\State $D_i \gets \mathcal{G}(\tilde{z})$ \Comment{Generate image sequence}

\State $\mathcal{R} \gets \emptyset$ \Comment{Initialize explanation set}
\For{$i = 1$ to $n$}
    \State $r_i \gets \pi_{\theta^\prime}(D_i, \mathcal{P})$ 
    \State $\mathcal{R} \gets \mathcal{R} \cup \{r_i\}$
\EndFor
\State $s_i = \frac{1}{n - 1} \sum_{ i \neq j} \textsc{Sim}(r_i, r_j)$
\State $\overline{s} = \frac{1}{n\cdot(n-1)} {\sum}_{i=1}^{n}{\sum}_{j=1, i \neq j}^n \textsc{Sim}(r_i, r_j)$ \Comment{Compute certainty score}
\State $\hat{r} \gets \arg\max_{r_i \in \mathcal{R}} s_i$
\If{$\overline{s} \geq \epsilon$}
    \State \Return $\hat{r}$
\Else
    \State \Return ``No clear explanation''
\EndIf
\end{algorithmic}
\end{algorithm}

\subsection{Automatic In-context Prompt Generation}
\label{sec:prompt}
By leveraging these three common inductive biases identified in generative models, we can automatically generate prompts $\mathcal{P}$ that align with $\mathcal{F}$ within these three biases using in-context learning. 

As Algorithm~\ref{alg:latentexplainer} shows, our approach starts by extracting mathematical symbols from the formula $\mathcal{F}$ using the \texttt{ExtractSymbols} function (line 4). This function traverses the formula to identify the mathematical symbols.

Next, the algorithm initializes an empty dictionary \texttt{semantics} to store the semantic representations of these symbols (line 5). For each symbol, the pre-trained LLM $\pi_\theta$ extracts its semantic meaning based on few-shot examples $\mathcal{H}$ and the optional input symbol information $\mathcal{I}$ (lines 6-9).

Finally, the algorithm generates the prompt $\mathcal{P}$ using the formula $\mathcal{F}$, the few-shot examples $\mathcal{H}$, and the gathered semantics (line 10). This step-by-step reasoning process ensures that the generated prompts are contextually relevant and aligned with the underlying formulas, which could reduce hallucination and enhance model performance.

\subsection{Inductive-bias-guided Data Manipulation}
\label{sec:manipulate}
First, identify the relevant formulas within the input inductive bias formulas with regard to a specific latent variable $z_i$ to be explained. Then, combine the identified relevant formulas with the inductive bias prompt $\mathcal{P}$ obtained from Section~\ref{sec:prompt} and utilize an LLM as a coding agent to generate prompts that specify the modifications needed in the generative model's decoder code (e.g., adjusting indices through the subscripts of the latent variables or adding properties according to extra information about the symbol $\mathcal{I}$). The coding agent then tests and refines the generated code to effectively perturb $z_i$. Through a series of perturbations on \( z_i \), a sequence of generated images can be obtained, capturing the variations in \( z_i \) and reflecting its influence on the generated data. Finally, all image sequences generated from the relevant formulas are aggregated to generate explanations about $z_i$. The implementation details are in Section~\ref{appendix:perturbation}.

\subsection{Uncertainty-aware Explanations}
\label{sec:uncertainty}
Uncertainty-aware methods can be applied to large language model responses \cite{lin2023generating} and image explanations \cite{zhao2024due}. To measure the uncertainty of the responses from \textit{GPT-4o}, we sampled $n$ times from the \textit{GPT-4o} to generate the responses $\mathcal{R}=\{r_1,r_2,r_3,...,r_n\}.$ The certainty score of the explanation is the average pairwise cosine similarity of the responses $\mathcal{R}$: $ 1/C\cdot{\sum}_{i=1}^{n}{\sum}_{j=1, i \neq j}^n\text{sim}(r_i,r_j)$
where $C = n\cdot(n-1)$. Our final explanation $\hat{r}$ is the response that has the highest pairwise cosine similarity with other responses if the latent variable is interpretable. Otherwise, $\hat{r}$ will be ``No clear explanation". The implementation details are in Section~\ref{appendix:uncertainty}.

\section{Experiment}
\label{sec:experiment}

The experiments are conducted on a 64-bit machine with 24-core Intel 13th Gen Core i9-13900K @ 5.80GHz, 32GB memory and NVIDIA GeForce RTX 4090. We set \textit{GPT-4o} at $temperature = 1$ and $top\_p = 1.$

\subsection{Dataset}

We utilized five datasets to evaluate the performance of different generative models under three inductive biases: CelebA-HQ \cite{karras2018progressive}, AFHQ \cite{choi2020stargan}, LSUN-Church \cite{yu15lsun} for the unconditional and conditional diffusion models, and CelebA-HQ \cite{karras2018progressive}, 3DShapes \cite{3dshapes18}, and dSprites \cite{dsprites17} for the VAE models. The \textbf{CelebA-HQ} dataset is a high-quality version of the CelebA dataset, consisting of 30K images of celebrities, divided into 28K for training and 2K for testing. The \textbf{LSUN-Church} dataset contains large-scale images of church buildings. The \textbf{AFHQ} dataset includes high-quality images of animals, divided into three categories: cats, dogs, and wild animals. \textbf{3DShapes} is a synthetic dataset contains images of 3D shapes with six factors of variation: floor hue, wall hue, object hue, object scale, object shape, and wall orientation, divided into 384K for training and 96K for testing. \textbf{dSprites} consists of 2D shapes (hearts, squares, ellipses) generated with five factors of variation: shape, scale, orientation, position X, and position Y, divided into 516K for training and 221K for testing.

\begin{table*}[h]
    \centering
    \caption{Quantitative results for diffusion models across datasets. B represents BLEU@4, R represents ROUGE-L, S represents SPICE, BS represents BERTScore, and BAS represents BARTScore.}
    \resizebox{\textwidth}{!}{
    \begin{tabular}{llcccccccccccccccccc}
        \toprule
        \multirow{2}{*}{\textbf{Model}} & \multirow{2}{*}{\textbf{Method}} & \multicolumn{5}{c}{\textbf{AFHQ}} & \multicolumn{5}{c}{\textbf{LSUN-Church}} & \multicolumn{5}{c}{\textbf{CelebA-HQ}} \\
       \cmidrule(lr){3-7} \cmidrule(lr){8-12} \cmidrule(lr){13-17}
         &  & $\textbf{B} \uparrow$ & $\textbf{R} \uparrow$ & $\textbf{S} \uparrow$ & $\textbf{BS} \uparrow$ & $\textbf{BAS} \uparrow$
        & $\textbf{B} \uparrow$ & $\textbf{R} \uparrow$ & $\textbf{S} \uparrow$ & $\textbf{BS} \uparrow$ & $\textbf{BAS}\uparrow$
         & $\textbf{B}\uparrow$ & $\textbf{R}\uparrow$  & $\textbf{S}\uparrow$ & $\textbf{BS}\uparrow$ & $\textbf{BAS} \uparrow$ \\
        \midrule
        \multirow{6}{*}{\parbox{2.5cm}{DDPM \\(Disentanglement Bias)}}
        & Gemini 1.5 Pro &2.87 &23.18 &11.46&88.27&-3.35 & 1.64&20.75&6.92&87.22&-3.51 & 1.80&22.18&8.45&87.96&-3.11 \\
        & \; + \textit{\textbf{LatentExplainer}} & \textbf{3.15} & \textbf{25.62} & \textbf{13.36} & \textbf{88.63} & \textbf{-3.30}
& \textbf{5.71} & \textbf{26.71} & \textbf{10.31} & \textbf{87.95} & \textbf{-3.46}
& \textbf{4.97} & \textbf{26.74} & \textbf{9.64} & \textbf{88.72} & \textbf{-3.08} \\
\cmidrule(lr){2-17}

& Claude 3.5 Sonnet&4.16&22.25&13.34&88.02&-3.28 & 7.76&28.02&14.14&88.65&-3.36 & 8.05&25.60&12.11&88.46&-3.10 \\
        & \; + \textit{\textbf{LatentExplainer}} & \textbf{12.52} & \textbf{31.68} & \textbf{20.05} & \textbf{89.75} & \textbf{-3.06}
& \textbf{12.92} & \textbf{32.55} & \textbf{19.52} & \textbf{89.70} & \textbf{-3.23}
& \textbf{14.14} & \textbf{34.07} & \textbf{14.80} & \textbf{90.06} & \textbf{-2.94} \\
\cmidrule(lr){2-17}

        & GPT-4o & 3.08	&14.83		&8.78	&87.55	&-3.39 & 5.52	&23.06		&11.23	&88.47	&-3.41 & 0.11	&3.54	&1.22	&85.97	&-3.16 \\
         & \; + \textit{\textbf{LatentExplainer}} & \textbf{25.91}	&\textbf{37.84}		&\textbf{29.87}	&\textbf{91.48}	&\textbf{-2.97} &\textbf{30.92}	&\textbf{44.21}		&\textbf{29.23}	&\textbf{91.80}	&\textbf{-3.06} &\textbf{18.49}	&\textbf{35.27}	&\textbf{18.81}	&\textbf{90.30}	&\textbf{-2.90}\\
        \midrule
        \multirow{6}{*}{\parbox{2.5cm}{Stable Diffusion \\(Conditional Bias)}}
        & Gemini 1.5 Pro &0.00&20.36&10.95&88.74&-3.34 & \textbf{0.00}&20.24&8.98&88.08&-3.56 & 0.00&18.93&7.48&87.83&-3.16 \\
        & \; + \textbf{LatentExplainer} & \textbf{3.29} & \textbf{21.52} & \textbf{15.21} & \textbf{89.09} & \textbf{-3.29}
& \textbf{0.00} & \textbf{21.49} & \textbf{9.57} & \textbf{88.09} & \textbf{-3.53}
& \textbf{1.89} & \textbf{21.22} & \textbf{8.60} & \textbf{88.01} & \textbf{-3.09} \\
\cmidrule(lr){2-17}
& Claude 3.5 Sonnet&6.57&30.62&17.80&89.53&-3.31 & 2.37&26.95&13.10&88.74&-3.55 & \textbf{4.69}&26.76&13.52&89.27&-3.15 \\
& \; + \textit{\textbf{LatentExplainer}} & \textbf{7.69} & \textbf{31.84} & \textbf{18.27} & \textbf{89.85} & \textbf{-3.28}
& \textbf{2.69} & \textbf{28.88} & \textbf{14.14} & \textbf{89.01} & \textbf{-3.46}
& 4.02 & \textbf{28.68} & \textbf{15.22} & \textbf{89.61} & \textbf{-3.08} \\
\cmidrule(lr){2-17}
        & GPT-4o & 7.61	&26.32	&17.19	&90.04	&-3.32 &7.80	&26.73	&14.28	&89.44	&-3.46 &5.79	&23.89	&12.68	&89.63	&-3.04 \\
        & \; + \textit{\textbf{LatentExplainer}} & \textbf{13.27}	&\textbf{29.98}	&\textbf{20.08}	&\textbf{90.17}	&\textbf{-3.19}
&\textbf{18.33}	&\textbf{38.14}		&\textbf{25.01}	&\textbf{90.99}	&\textbf{-3.23}
&\textbf{18.50}	&\textbf{40.85}	&\textbf{23.90}	&\textbf{91.75}	&\textbf{-2.82}\\

        \bottomrule
    \end{tabular}
    }
    \label{appendix:quantitative_diffusion}
\end{table*}

\begin{table*}[h]
    \centering
    \caption{Quantitative results for VAE models across datasets. B represents BLEU@4, R represents ROUGE-L, S represents SPICE, BS represents BERTScore, and BAS represents BARTScore.}
    \resizebox{\textwidth}{!}{%
    \begin{tabular}{llcccccccccccccccccc}
        \toprule
        \multirow{2}{*}{\textbf{Model}} & \multirow{2}{*}{\textbf{Method}} & \multicolumn{5}{c}{\textbf{3DShapes}} & \multicolumn{5}{c}{\textbf{CelebA-HQ}} & \multicolumn{5}{c}{\textbf{dSprites}} \\
        \cmidrule(lr){3-7} \cmidrule(lr){8-12} \cmidrule(lr){13-17}
         &  & $\textbf{B} \uparrow$ & $\textbf{R} \uparrow$ & $\textbf{S} \uparrow$ & $\textbf{BS} \uparrow$ & $\textbf{BAS} \uparrow$
        & $\textbf{B} \uparrow$ & $\textbf{R} \uparrow$& $\textbf{S} \uparrow$& $\textbf{BS} \uparrow$& $\textbf{BAS}\uparrow$
         & $\textbf{B}\uparrow$ & $\textbf{R}\uparrow$  & $\textbf{S}\uparrow$& $\textbf{BS}\uparrow$ & $\textbf{BAS}\uparrow$ \\
        \midrule
        \multirow{6}{*}{\parbox{2.5cm}{$\beta$-TCVAE \\(Disentanglement Bias)}}
        &Gemini 1.5 Pro
&2.73&26.45&5.56&88.83&-3.15
&0.00&17.28&4.86&87.00&-3.27
&0.00&15.39&6.11&86.61&-3.25 \\
&  \; + \textit{\textbf{LatentExplainer}}
& \textbf{3.11} & \textbf{28.98} & \textbf{8.07} & \textbf{88.94} & \textbf{-3.14}
& \textbf{2.19} & \textbf{31.21} & \textbf{10.70} & \textbf{88.98} & \textbf{-3.18}
& \textbf{1.54} & \textbf{17.29} & \textbf{7.61} & \textbf{86.98} & \textbf{-3.22} \\
\cmidrule(lr){2-17}

&Claude 3.5 Sonnet
&11.02&30.95&14.13&89.11&-3.01
&9.54&30.96&11.11&88.76&-3.14
&3.65&22.32&11.91&88.49&-3.05\\
&  \;  + \textit{\textbf{LatentExplainer}}
& \textbf{18.09} & \textbf{33.69}& \textbf{14.26} & \textbf{89.36} & \textbf{-2.92}
& \textbf{12.73} & \textbf{33.27} & \textbf{11.21} & \textbf{89.50} & \textbf{-3.05}
& \textbf{12.20} & \textbf{35.87} & \textbf{19.29} & \textbf{90.58} & \textbf{-2.95} \\
\cmidrule(lr){2-17}

        & GPT-4o 
&5.51	&29.77	&7.92	&89.18	&-3.13
&11.09	&37.14	&17.24	&90.55	&-3.07
&0.00	&16.19	&7.04	&87.17	&-3.22\\
         & \;  + \textit{\textbf{LatentExplainer}} 
&\textbf{25.40}	&\textbf{49.06}		&\textbf{22.78}	&\textbf{91.90}	&\textbf{-2.75}
&\textbf{29.93}	&\textbf{55.68}		&\textbf{30.17}	&\textbf{93.55}	&\textbf{-2.77}
& \textbf{12.55}	&\textbf{37.30}	&\textbf{21.69}	&\textbf{90.28}	&\textbf{-2.87}\\
        \midrule
        \multirow{6}{*}{\parbox{2.5cm}{CSVAE \\(Combination Bias)}}  
        &Gemini 1.5 Pro
&3.11&22.14&5.39&88.33&-2.99
&1.21&16.52&5.28&88.25&\textbf{-3.21}
&1.18&23.15&5.83&88.24&-3.15 \\
&  \; + \textit{\textbf{LatentExplainer}}
& \textbf{6.22} & \textbf{25.59} & \textbf{5.69} & \textbf{88.98} & \textbf{-2.97}
& \textbf{3.88} & \textbf{24.17} & \textbf{7.94} & \textbf{88.72} & \textbf{-3.21}
& \textbf{1.33} & \textbf{26.43} & \textbf{5.88} & \textbf{89.18} & \textbf{-3.08} \\
\cmidrule(lr){2-17}

&Claude 3.5 Sonnet
&4.55&24.30&9.14&87.08&-3.01
&9.43&28.77&18.31&89.50&-3.11
&7.04&24.16&14.27&88.45&-3.08\\

&  \; + \textit{\textbf{LatentExplainer}}
& \textbf{15.24} & \textbf{25.57} & \textbf{19.82} & \textbf{88.72} & \textbf{-2.87}
& \textbf{19.92} & \textbf{28.85} & \textbf{18.97} & \textbf{89.55} & \textbf{-2.97}
& \textbf{17.75} & \textbf{30.87} & \textbf{22.14} & \textbf{89.58} & \textbf{-2.94} \\
\cmidrule(lr){2-17}

        & GPT-4o 
&6.93	&25.07	&9.70	&89.42	&-2.87
&10.44	&37.42	&15.06	&90.53	&-3.01 
& 0.00	&21.61	&6.99	&87.19	&-3.17\\
        &  \; + \textit{\textbf{LatentExplainer}} 
&\textbf{36.18}	&\textbf{43.58}	&\textbf{36.75}	&\textbf{91.72}	&\textbf{-2.52}
&\textbf{25.34}	&\textbf{45.18}		&\textbf{26.96}	&\textbf{92.09}	&\textbf{-2.81}
& \textbf{16.03}	&\textbf{35.85}	&\textbf{21.82}	&\textbf{90.05}	&\textbf{-2.90}\\

        \midrule
        \multirow{6}{*}{\parbox{2.5cm}{CSVAE \\(Conditional Bias)}}
        &Gemini 1.5 Pro
&8.61&26.41&7.11&87.95&-3.20
&3.48&19.77&10.41&87.92&-3.12
&0.00&17.27&8.65&86.80&-3.30 \\
&  \; + \textit{\textbf{LatentExplainer}}
& \textbf{12.12} & \textbf{35.80} & \textbf{9.38} & \textbf{90.14} & \textbf{-3.08}
& \textbf{3.63} & \textbf{21.26} & \textbf{10.65} & \textbf{87.95} & \textbf{-3.05}
& \textbf{2.62} & \textbf{22.99} & \textbf{10.53} & \textbf{88.26} & \textbf{-3.18} \\
\cmidrule(lr){2-17}
&Claude 3.5 Sonnet
&5.36&26.69&15.28&88.75&-3.04
&13.83&34.38&18.15&90.36&-2.98
&8.16&25.09&17.70&88.18&-3.11\\
&  \; + \textit{\textbf{LatentExplainer}}
& \textbf{8.42} & \textbf{31.32} & \textbf{18.34} & \textbf{88.82} & \textbf{-2.93}
& \textbf{14.53} & \textbf{34.54} & \textbf{19.58} & \textbf{90.70} & \textbf{-2.90}
& \textbf{8.42} & \textbf{27.99} & \textbf{17.94} & \textbf{88.34} & \textbf{-3.04} \\
\cmidrule(lr){2-17}

        & GPT-4o 
&10.97	&30.55	&8.83	&90.10	&-3.02
&8.46	&28.78	&15.78	&89.62	&-2.98 
& 0.00	&9.62	&1.03	&85.74	&-3.34\\
        & \; + \textit{\textbf{LatentExplainer}} 
&\textbf{25.71}	&\textbf{40.00}	&\textbf{20.67}	&\textbf{90.88}	&\textbf{-2.79}
&\textbf{24.21}	&\textbf{42.88}	&\textbf{19.61}	&\textbf{91.21}	&\textbf{-2.79}
& \textbf{21.17}	&\textbf{34.99}	&\textbf{20.71}	&\textbf{89.49}	&\textbf{-3.08}\\

        \bottomrule
    \end{tabular}
    }
    \label{appendix:quantitative_vae}
    
\end{table*}

\begin{table*}[h]
    \centering
    \caption{Ablation results for diffusion models across datasets on GPT-4o. B represents BLEU@4, R represents ROUGE-L, S represents SPICE, BS represents BERTScore, and BAS represents BARTScore.}
    \resizebox{\textwidth}{!}{
    \begin{tabular}{llcccccccccccccccccc}
        \toprule
        \multirow{2}{*}{\textbf{Model}} & \multirow{2}{*}{\textbf{Method}} & \multicolumn{5}{c}{\textbf{AFHQ}} & \multicolumn{5}{c}{\textbf{LSUN-Church}} & \multicolumn{5}{c}{\textbf{CelebA-HQ}} \\
       \cmidrule(lr){3-7} \cmidrule(lr){8-12} \cmidrule(lr){13-17}
         &  & $\textbf{B} \uparrow$ & $\textbf{R} \uparrow$ & $\textbf{S} \uparrow$ & $\textbf{BS} \uparrow$ & $\textbf{BAS} \uparrow$
        & $\textbf{B} \uparrow$ & $\textbf{R} \uparrow$ & $\textbf{S} \uparrow$ & $\textbf{BS} \uparrow$ & $\textbf{BAS}\uparrow$
         & $\textbf{B}\uparrow$ & $\textbf{R}\uparrow$  & $\textbf{S}\uparrow$ & $\textbf{BS}\uparrow$ & $\textbf{BAS} \uparrow$ \\
        \midrule
        \multirow{4}{*}{\parbox{2.5cm}{DDPM \\(Disentanglement Bias)}}

        & GPT-4o & 3.08	&14.83		&8.78	&87.55	&-3.39 & 5.52	&23.06		&11.23	&88.47	&-3.41 & 0.11	&3.54	&1.22	&85.97	&-3.16 \\
         & \; + LatentExplainer w/o IB & 2.14	&11.49	&7.82	&87.17	&-3.41 &8.78	&27.01	&14.08	&88.85	&-3.37 &0.19	&4.17		&1.80	&85.88	&-3.16\\
         & \; + LatentExplainer w/o UQ & 16.12	&31.57	&21.70	&89.99	&-3.12 &21.19	&37.03		&23.65	&90.49	&-3.19 &13.11	&27.38		&14.03	&89.15	&-2.92\\
         &  \; + \textit{\textbf{LatentExplainer}} & \textbf{25.91}	&\textbf{37.84}		&\textbf{29.87}	&\textbf{91.48}	&\textbf{-2.97} &\textbf{30.92}	&\textbf{44.21}		&\textbf{29.23}	&\textbf{91.80}	&\textbf{-3.06} &\textbf{18.49}	&\textbf{35.27}	&\textbf{18.81}	&\textbf{90.30}	&\textbf{-2.90}\\
        \midrule
        \multirow{4}{*}{\parbox{2.5cm}{Stable Diffusion \\(Conditional Bias)}}
              & GPT-4o & 7.61	&26.32	&17.19	&90.04	&-3.32 &7.80	&26.73	&14.28	&89.44	&-3.46 &5.79	&23.89	&12.68	&89.63	&-3.04 \\
        & \; + LatentExplainer w/o IB & 9.12	&26.59	&16.28	&89.77	&-3.29 &10.07	&28.31	&15.85&	89.92	&-3.42 &5.87	&24.06	&12.17	&89.15	&-3.05\\
        & \; + LatentExplainer w/o UQ & 12.36	&27.73	&20.01	&90.03&	\textbf{-3.18}
&17.19	&35.33		&21.76	&90.73	&-3.31
&15.01	&35.43	&\textbf{23.99}	&91.07	&-2.87\\
        &  \; + \textit{\textbf{LatentExplainer}} & \textbf{13.27}	&\textbf{29.98}	&\textbf{20.08}	&\textbf{90.17}	&-3.19
&\textbf{18.33}	&\textbf{38.14}		&\textbf{25.01}	&\textbf{90.99}	&\textbf{-3.23}
&\textbf{18.50}	&\textbf{40.85}	&23.90	&\textbf{91.75}	&\textbf{-2.82}\\

        \bottomrule
    \end{tabular}
    }
    \label{ablation:quantitative_diffusion}
\end{table*}

\begin{table*}[h]
    \centering
    \caption{Ablation results for VAE models across datasets on GPT-4o. B represents BLEU@4, R represents ROUGE-L, S represents SPICE, BS represents BERTScore, and BAS represents BARTScore.}
    \resizebox{\textwidth}{!}{%
    \begin{tabular}{llcccccccccccccccccc}
        \toprule
        \multirow{2}{*}{\textbf{Model}} & \multirow{2}{*}{\textbf{Method}} & \multicolumn{5}{c}{\textbf{3DShapes}} & \multicolumn{5}{c}{\textbf{CelebA-HQ}} & \multicolumn{5}{c}{\textbf{dSprites}} \\
        \cmidrule(lr){3-7} \cmidrule(lr){8-12} \cmidrule(lr){13-17}
         &  & $\textbf{B} \uparrow$ & $\textbf{R} \uparrow$ & $\textbf{S} \uparrow$ & $\textbf{BS} \uparrow$ & $\textbf{BAS} \uparrow$
        & $\textbf{B} \uparrow$ & $\textbf{R} \uparrow$& $\textbf{S} \uparrow$& $\textbf{BS} \uparrow$& $\textbf{BAS}\uparrow$
         & $\textbf{B}\uparrow$ & $\textbf{R}\uparrow$  & $\textbf{S}\uparrow$& $\textbf{BS}\uparrow$ & $\textbf{BAS}\uparrow$ \\
        \midrule
        \multirow{4}{*}{\parbox{2.5cm}{$\beta$-TCVAE \\(Disentanglement Bias)}}
        & GPT-4o 
&5.51	&29.77	&7.92	&89.18	&-3.13
&11.09	&37.14	&17.24	&90.55	&-3.07
&0.00	&16.19	&7.04	&87.17	&-3.22\\
         & \; + LatentExplainer w/o IB  
&5.41	&31.66	&10.07	&89.44	&-3.13
&6.14	&32.73	&14.04&	89.92	&-3.11
&0.00	&18.73	&7.54	&87.42&	-3.20\\
         & \; + LatentExplainer w/o UQ 
&16.99	&37.37	&\textbf{22.89}	&90.83	&-2.80
&21.95	&48.84	&22.60	&92.09	&-2.88
& \textbf{17.16}	&37.19	&\textbf{22.40}	&90.18	&-2.89\\
         &  \; + \textit{\textbf{LatentExplainer}} 
&\textbf{25.40}	&\textbf{49.06}		&22.78	&\textbf{91.90}	&\textbf{-2.75}
&\textbf{29.93}	&\textbf{55.68}		&\textbf{30.17}	&\textbf{93.55}	&\textbf{-2.77}
& 12.55	&\textbf{37.30}	&21.69	&\textbf{90.28}	&\textbf{-2.87}\\
        \midrule
        \multirow{4}{*}{\parbox{2.5cm}{CSVAE \\(Combination Bias)}}  
        & GPT-4o 
&6.93	&25.07	&9.70	&89.42	&-2.87
&10.44	&37.42	&15.06	&90.53	&-3.01 
& 0.00	&21.61	&6.99	&87.19	&-3.17\\
        & \; + LatentExplainer w/o IB 
&14.22	&28.57	&11.79	&90.02	&-2.85
&11.96	&37.02	&17.49	&90.50	&-3.01
& 0.00	&16.36	&7.26	&86.96	&-3.24\\
        & \; + LatentExplainer w/o UQ 
&34.55	&39.62	&30.30	&90.83	&-2.62
&14.08	&38.68	&22.49	&91.11	&-2.87
& 0.00	&28.49	&15.65	&89.78	&-2.91\\
        &  \; + \textit{\textbf{LatentExplainer}} 
&\textbf{36.18}	&\textbf{43.58}	&\textbf{36.75}	&\textbf{91.72}	&\textbf{-2.52}
&\textbf{25.34}	&\textbf{45.18}		&\textbf{26.96}	&\textbf{92.09}	&\textbf{-2.81}
& \textbf{16.03}	&\textbf{35.85}	&\textbf{21.82}	&\textbf{90.05}	&\textbf{-2.90}\\

        \midrule
        \multirow{4}{*}{\parbox{2.5cm}{CSVAE \\(Conditional Bias)}}
       
        & GPT-4o 
&10.97	&30.55	&8.83	&90.10	&-3.02
&8.46	&28.78	&15.78	&89.62	&-2.98 
& 0.00	&9.62	&1.03	&85.74	&-3.34\\
        & \; + LatentExplainer w/o IB 
&19.28	&39.90	&15.74	&90.87	&-2.88
&8.35	&28.75	&12.11	&89.60	&-2.97
& 0.00	&11.38	&5.26	&86.47	&-3.29\\
        & \; + LatentExplainer w/o UQ  
&16.73	&32.28	&16.19	&89.84	&-2.86
&13.35	&37.20	&\textbf{20.05}	&89.89	&-2.92
&5.39	&19.20	&6.36	&86.94	&-3.26\\
        &  \; + \textit{\textbf{LatentExplainer}} 
&\textbf{25.71}	&\textbf{40.00}	&\textbf{20.67}	&\textbf{90.88}	&\textbf{-2.79}
&\textbf{24.21}	&\textbf{42.88}	&19.61	&\textbf{91.21}	&\textbf{-2.79}
& \textbf{21.17}	&\textbf{34.99}	&\textbf{20.71}	&\textbf{89.49}	&\textbf{-3.08}\\

        \bottomrule
    \end{tabular}
    }
    \label{ablation:quantitative_vae}
    
\end{table*}

\subsection{Models and Baselines}

 Our evaluation benchmarks our proposed \textit{LatentExplainer} framework against three state-of-the-art multimodal models with strong vision-language reasoning capabilities: GPT-4o~\cite{openai2024gpt4o}, Gemini 1.5 Pro~\cite{team2024gemini}, and Claude 3.5 Sonnet~\cite{claude35sonnet2024}. We employ GPT-4o, Gemini 1.5 Pro, and Claude 3.5 as a zero-shot baseline, comparing it with the addition of our Latent Exlainer with both the inductive bias prompt and uncertainty quantification included.

\subsection{Generative Models under Inductive Biases }

We explore the latent space in generative models that satisfy the aforementioned three types of inductive biases. For each type, we present the relevant generative models that align with the corresponding inductive bias: (1) Disentanglement Bias: $\beta$-TCVAE \cite{chen2018isolating} explicitly penalizes the total correlation of the latent variables to disentangle the latent variables. Denoising Diffusion Probabilistic Model (DDPM) \cite{ho2020denoising} adds Gaussian noise independently at each timestep in the forward process and eventually transforms into pure Gaussian noise, in which the covariance matrix is diagonal. This assumes the latent factors are independent; (2) Combination Bias: CSVAE \cite{klys2018learning} has two groups of latent variables $z$ and $w$, where $z$ and $w$ are uncorrelated and the latent variables within the group are correlated; (3) Conditional Bias: CSVAE also satisfies conditional bias because one group of latent variables $w$ is associated with the properties while the other group of latent variables $z$ minimizes the mutual information with the properties. Stable Diffusion \cite{rombach2022high} is a latent diffusion model to generate images conditioned on prompts.

\begin{comment}
We use Singular Value Decomposition(SVD) on the Jacobian of the mapping from latent space to feature space to find the orthogonal meaningful latent directions to learn the disentanglement in Denoising Diffusion Probabilistic Models(DDPM) \cite{ho2020denoising}; (2)
\end{comment}

\begin{figure*}[t]
    \centering
    \begin{minipage}{0.49\textwidth}
        \centering
        \includegraphics[width=\textwidth]{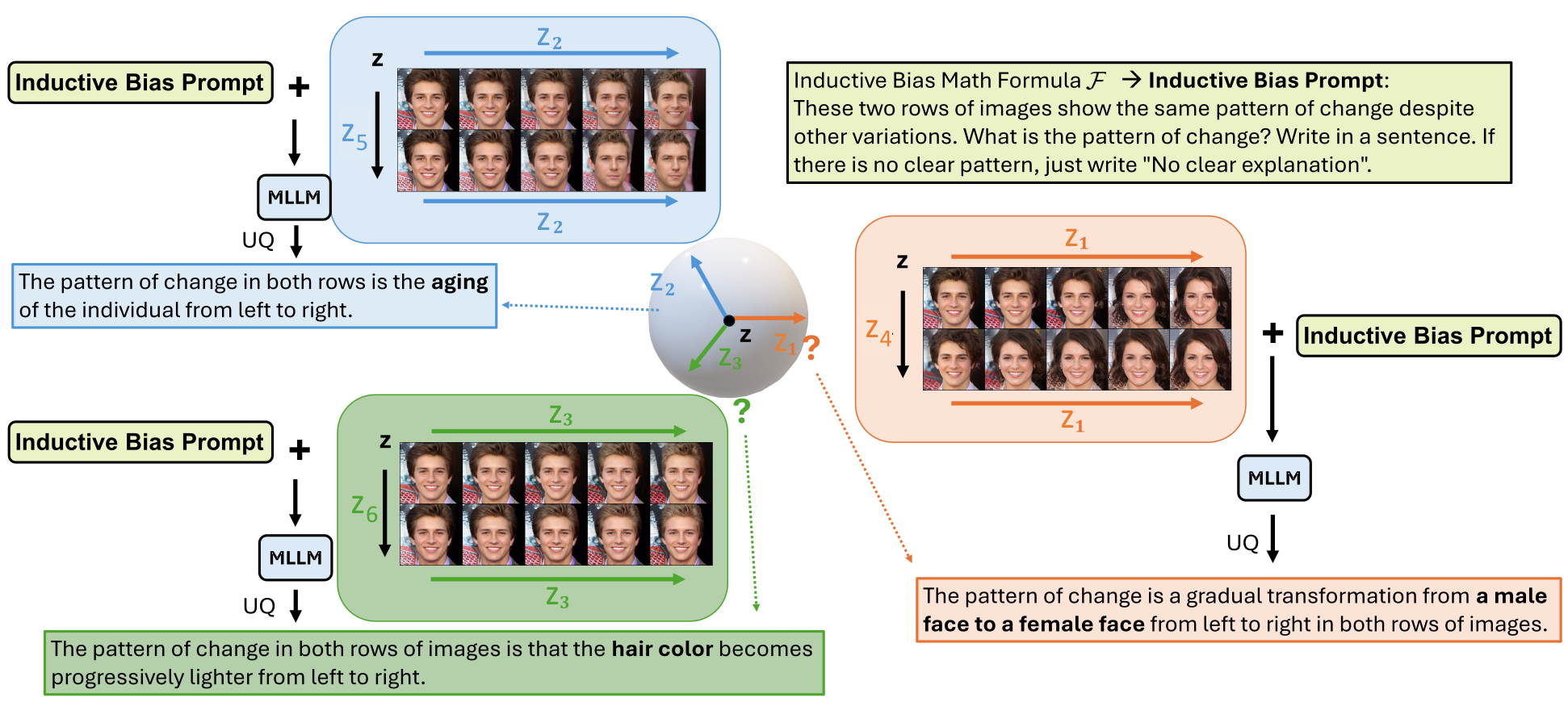}
        \caption{Visualization of the generated explanations with the inductive bias prompt for the disentanglement bias.}
        \label{fig:evaluation_disentanglement}
    \end{minipage}
    \hfill
    \begin{minipage}{0.49\textwidth}
        \centering
        \includegraphics[width=\textwidth]{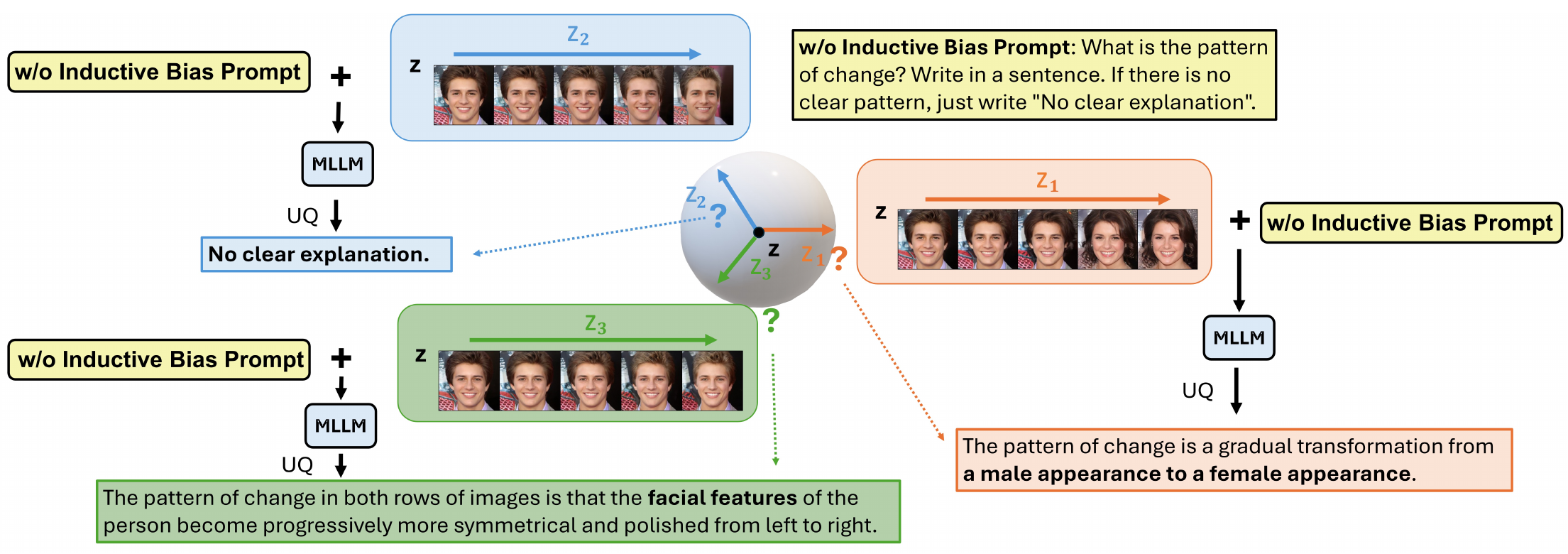}
        \caption{Visualization of the generated explanations w/o the inductive bias prompt for the disentanglement bias.}
        \label{fig:evaluation_disentanglement_wo}
    \end{minipage}
\end{figure*}

\begin{figure*}[t]
    \centering
    \begin{minipage}{0.49\textwidth}
        \centering
        \includegraphics[width=\textwidth]{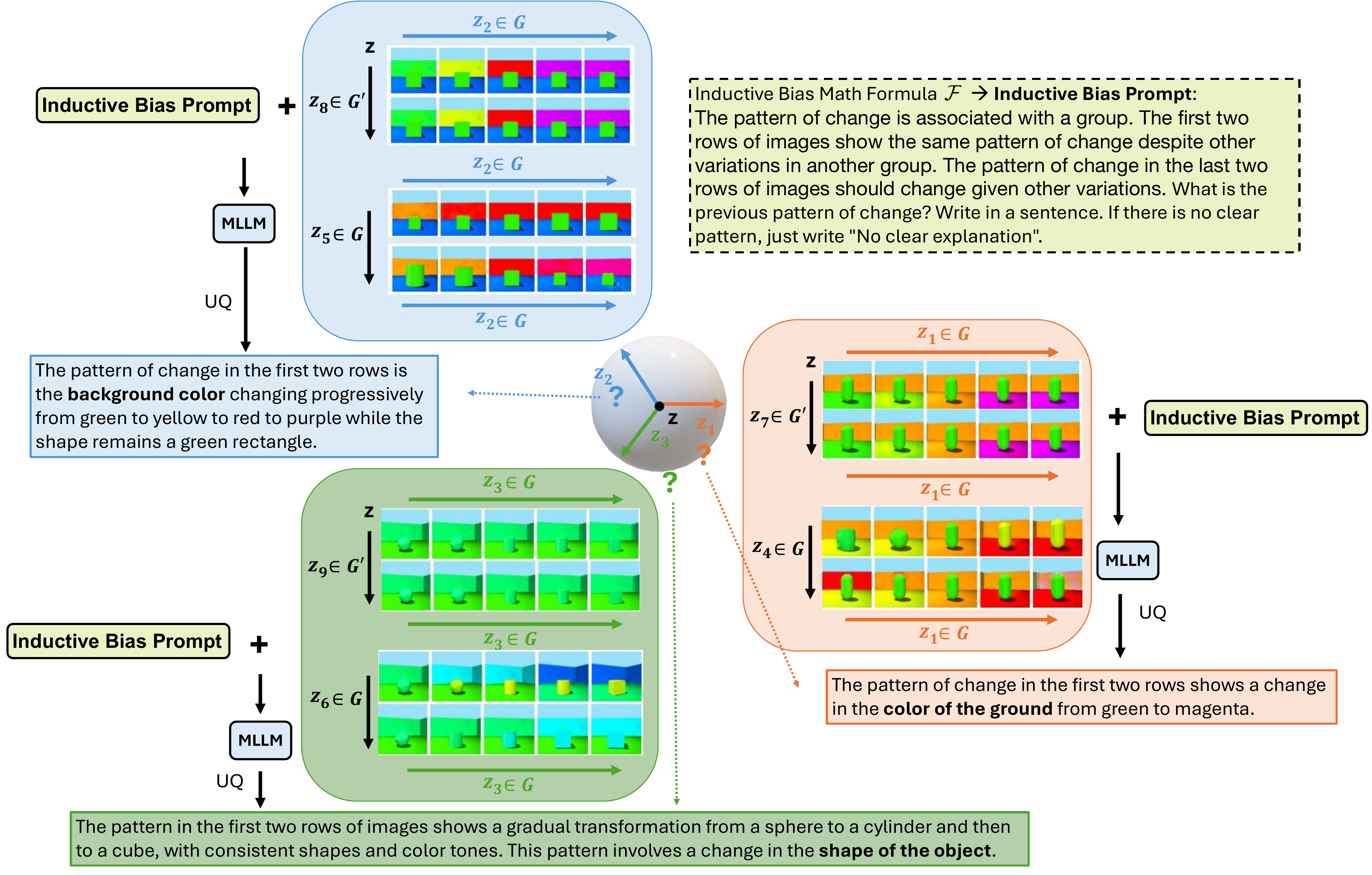}
\caption{Visualization of the generated explanations with the inductive bias prompt for the combination bias.}
\label{fig:evaluation_combination}
\end{minipage}
    \hfill
    \begin{minipage}{0.49\textwidth}
        \centering
        \includegraphics[width=\textwidth]{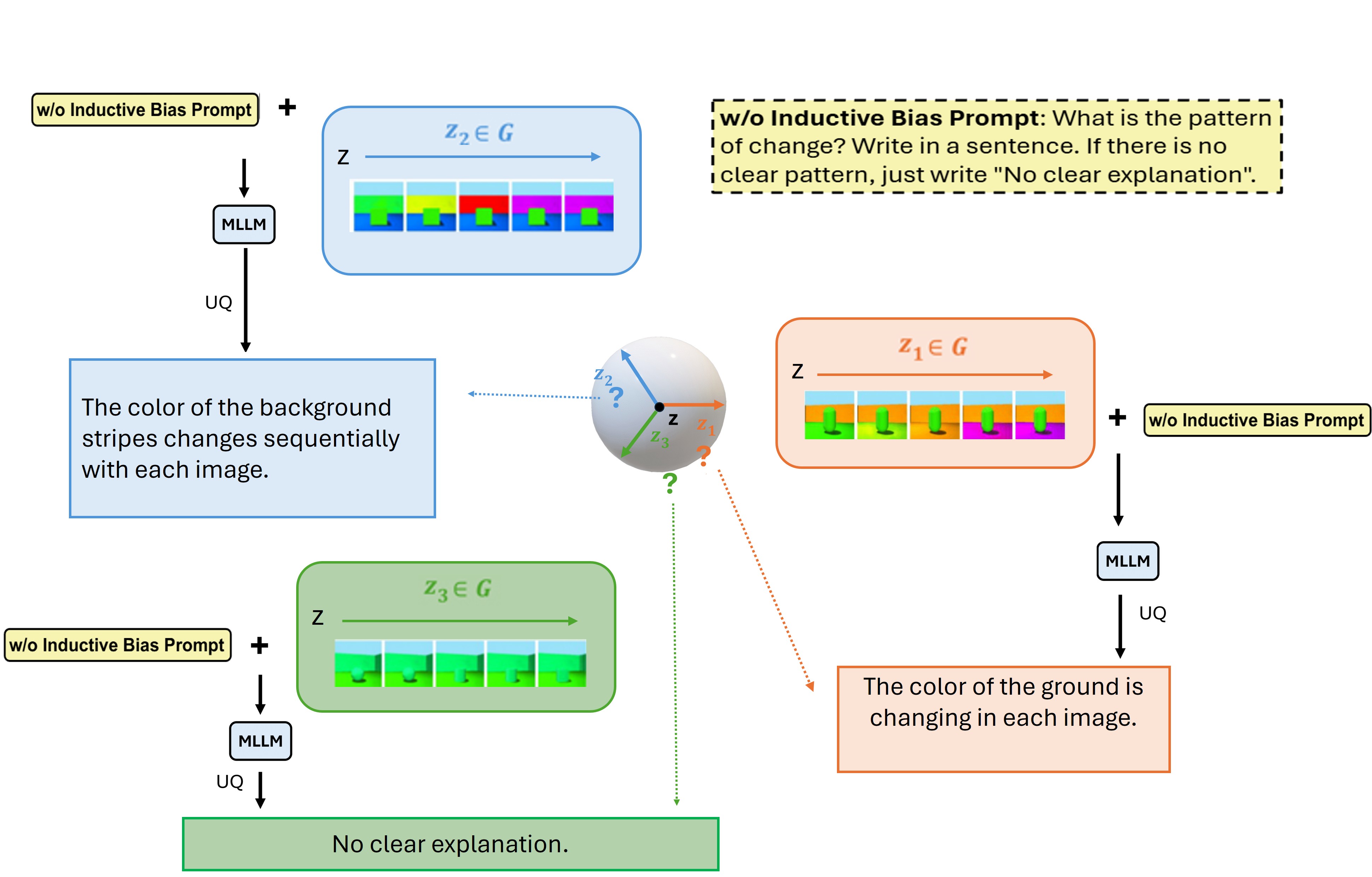}
\caption{Visualization of the generated explanations without the inductive bias prompt for the combination bias.}
\label{fig:evaluation_combination_wo}
\end{minipage}
\end{figure*}

\begin{figure*}[t]
    \centering
    \begin{minipage}{0.49\textwidth}
        \centering
        \includegraphics[width=\textwidth]{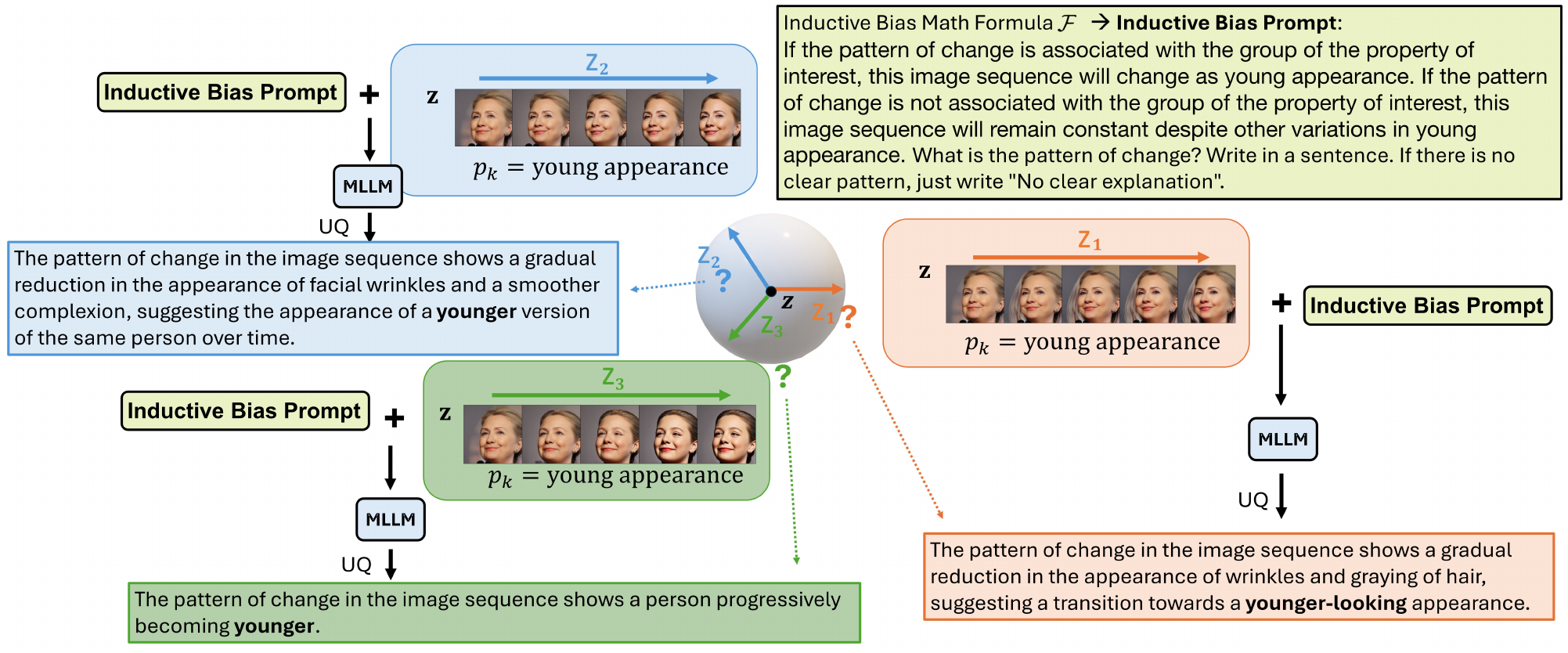}
\caption{Visualization of the generated explanations with the inductive bias prompt for the conditional bias.}
\label{fig:evaluation_conditional}
 \end{minipage}
    \hfill
    \begin{minipage}{0.49\textwidth}
        \centering
        \includegraphics[width=\textwidth]{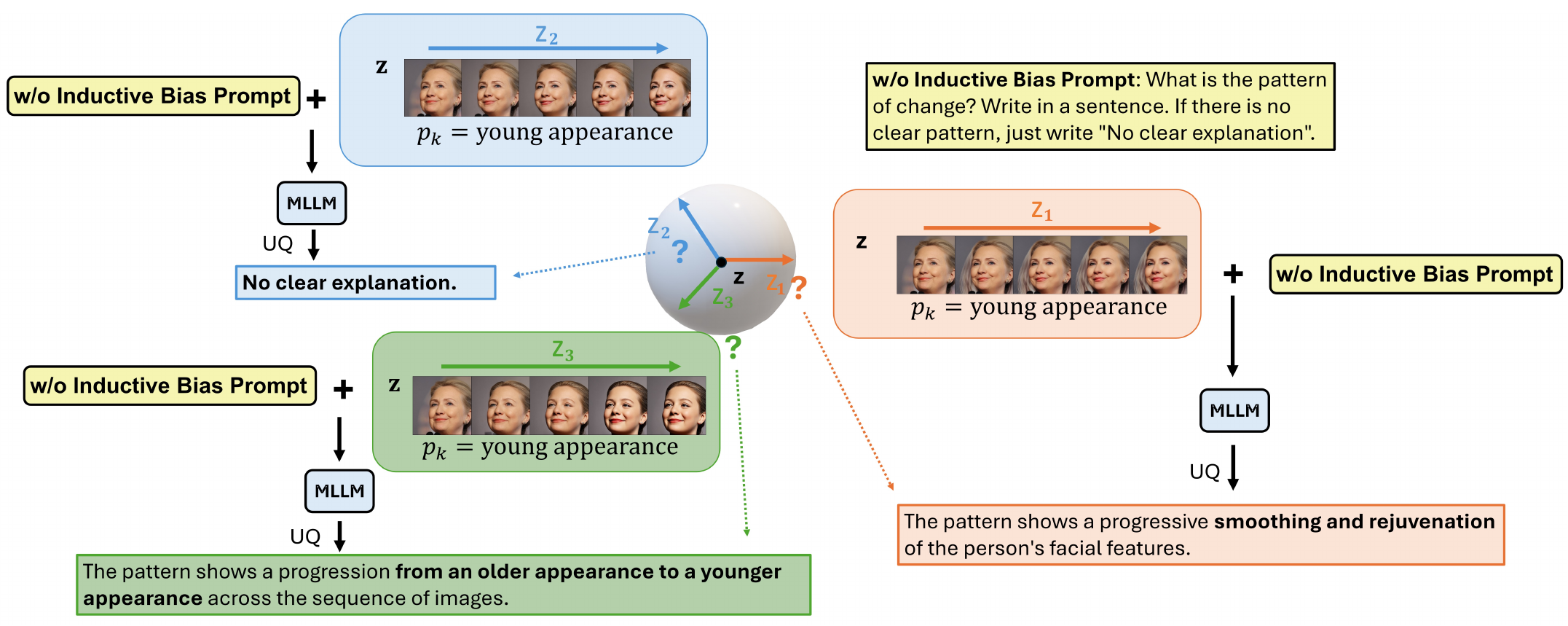}
\caption{Visualization of the generated explanations without the inductive bias prompt for the conditional bias.}
\label{fig:evaluation_conditional_wo}
    \end{minipage}
\end{figure*}

\vspace*{3cm}
\subsection{Latent Variable Perturbation Implementations}
\label{appendix:perturbation}
For VAE models, every time we target at each latent variable. The targeted latent variable is changed from -3 to 3, which is the 3 times the standard deviation of the standard Gaussian distribution of a latent variable, with a step size of 1.5. All the other latent variables are unchanged. 
The value of a specific latent variable among other latent variables can be determined by decomposing the inductive bias formulas. For diffusion models, we follow $\tilde{\mathbf{z}}=\mathbf{z}+\gamma\left[\mathcal{G}(\mathbf{z}+\mathbf{z_i})-\mathcal{G}(\mathbf{z})\right]$, where $\gamma$ is the strength and $\mathcal{G}$ is a diffusion decoder. We set the value of $\gamma$ to 0.1, 0.2, 0.3, 0.4, and 0.5 for DDPM, and to 1, 2, 3, 4, and 5 for Stable Diffusion. $\mathbf{z_i}$ is determined by decomposing the inductive bias formulas.

\subsection{Uncertainty-aware Implementations}
\label{appendix:uncertainty}
We follow \cite{zhu2024explaining} to denote the true label of the interpretability for the i-th latent variable $y_i$ as 1 if at least two of the three annotators can see a clear pattern in the generated images, otherwise we denote it as 0. The certainty score of the explanation for the i-th latent variable is $\bar s_i$. We adopt the Jaccard Index to measure similarity between the predicted label and the true label,
\begin{equation*}
\begin{split}
    & E(\varepsilon) = \texttt{Jaccard}(f(\bar s_i, \varepsilon), y_i),  \text{   where} \\
    & f(\bar s_i, \varepsilon) = \mathds{1}(\bar s_i \geq \varepsilon).
\end{split}
\end{equation*}

The threshold is then selected as the one with the maximum similarity with the true label,
\begin{equation*}
\varepsilon^* = \arg \max_{\varepsilon} E(\varepsilon).
\end{equation*}

By solving this equation across all datasets, we find the threshold $\varepsilon = 0.2617$. Since $\varepsilon$ is solved across all datasets, it reduces the dependence on any specific dataset. Our final output $\hat{r}$ is the response that has the highest mean pairwise cosine similarity with other responses if the certainty score is equal or greater than the threshold $\varepsilon$. Otherwise, $\hat{r}$ will be "no clear explanation". 

\subsection{Human Annotaions}
The ground-truth annotations of explanations are performed by three annotators from the United States and China. All annotators are students that had at least an undergraduate degree. Annotators were presented with the same images and the prompts as MLLMs and were asked to annotate the pattern of the images.  If there is no clear pattern, just write "No clear explanation". The annotations are then aggregated as references to calculate the automated evaluation metrics.

\subsection{Quantitative Analysis}

For the quantitative explanation evaluation, we use BLEU~\cite{papineni2002bleu}, ROUGE-L~\cite{lin-2004-rouge},  SPICE~\cite{anderson2016spice}, BERTScore\cite{zhang2019bertscore}, and BARTScore~\cite{yuan2021bartscore} as the automated metrics to assess the generated explanations. BLEU, and ROUGE-L are n-gram-based metrics that measure the overlap between generated and reference texts. SPICE compares scene graphs derived from the generated and reference texts. BERTScore and BARTScore utilize pre-trained transformer-based language models to compute contextual embeddings of the generated and reference texts. These metrics together provide a comprehensive assessment of both the lexical and semantic quality of the explanations. 

We evaluate the performance of Gemini 1.5 Pro, Claude 3.5 Sonnet and GPT-4o across all diffusion and vae models and datasets. Our proposed LatentExlainer method is a model-agnostic module and the inclusion of LatentExplainer consistently leads to substantial performance improvements on all MLLMs -- Gemini 1.5 Pro, Claude 3.5 Sonnet and GPT-4o. Across diffusion-based models in Table~\ref{appendix:quantitative_diffusion}, LatentExplainer delivers significant improvements across all datasets and metrics. Notably, when integrated with GPT-4o on CelebA-HQ, BLEU improves from 5.79 to 18.50, ROUGE-L from 23.89 to 40.85, and SPICE from 12.68 to 23.90, highlighting nearly 2x gains. Similar trends are observed with Claude 3.5 Sonnet for DDPM, where BLEU on AFHQ increases from 4.16 to 12.52, and ROUGE-L rises from 22.25 to 31.68, and SPICE from 13.34 to 20.05. Even for weaker baselines like Gemini 1.5 Pro, LatentExplainer consistently improves both lexical and semantic quality, demonstrating its effectiveness in enhancing explainability for latent variables.

For VAE-based models in Table~\ref{appendix:quantitative_vae}, LatentExplainer achieves even more striking gains. On the $\beta$-TCVAE model with GPT-4o, performance on 3DShapes sees BLEU increase from 5.51 to 25.40, and SPICE from 7.92 to 22.78. Across dSprites, the BLEU jump from 0.00 to 12.55 and ROUGE-L from 16.19 to 37.30 further demonstrate the utility of our approach. In summary, LatentExplainer consistently boosts the interpretability and textual quality of explanations for latent variable models across settings.

\subsection{Ablation Study}

To understand the impact of various components in our proposed \textit{LatentExplainer}, we conducted a comprehensive ablation study on GPT-4o across different models (DDPM, $\beta$-TCVAE, Stable Diffusion, and CSVAE) and inductive biases (disentanglement, combination, and conditional). Specifically, we compare the removal of inductive bias prompts (IB), the removal of uncertainty quantification (UQ), and the full model against the baseline GPT-4o. The results for each dataset are provided in Table~\ref{ablation:quantitative_diffusion} and Table~\ref{ablation:quantitative_vae}. Removing inductive bias prompts leads to a substantial drop in all generative models. Their consistent results demonstrate that inductive bias is the most important and necessary component when explaining the latent variables of generative models. The removal of uncertainty quantification also results in a slightly decreased performance in all generative models, indicating that uncertainty quantification is also effective, though not as critical as inductive bias prompts. The full model, which incorporates both inductive bias prompts and uncertainty quantification, achieves the highest overall performance, and outperforms all baselines across all models. This confirms the necessity of both inductive bias prompts and uncertainty quantification in our \textit{LatentExplainer} framework, demonstrating their significant contributions to improving explanatory performance across various generative models. It also shows that our framework can effectively design prompts for different inductive biases in generative models to improve the accuracy and reduce hallucination. 

\begin{figure}[t]
\begin{center}
\includegraphics[width=0.45\textwidth]{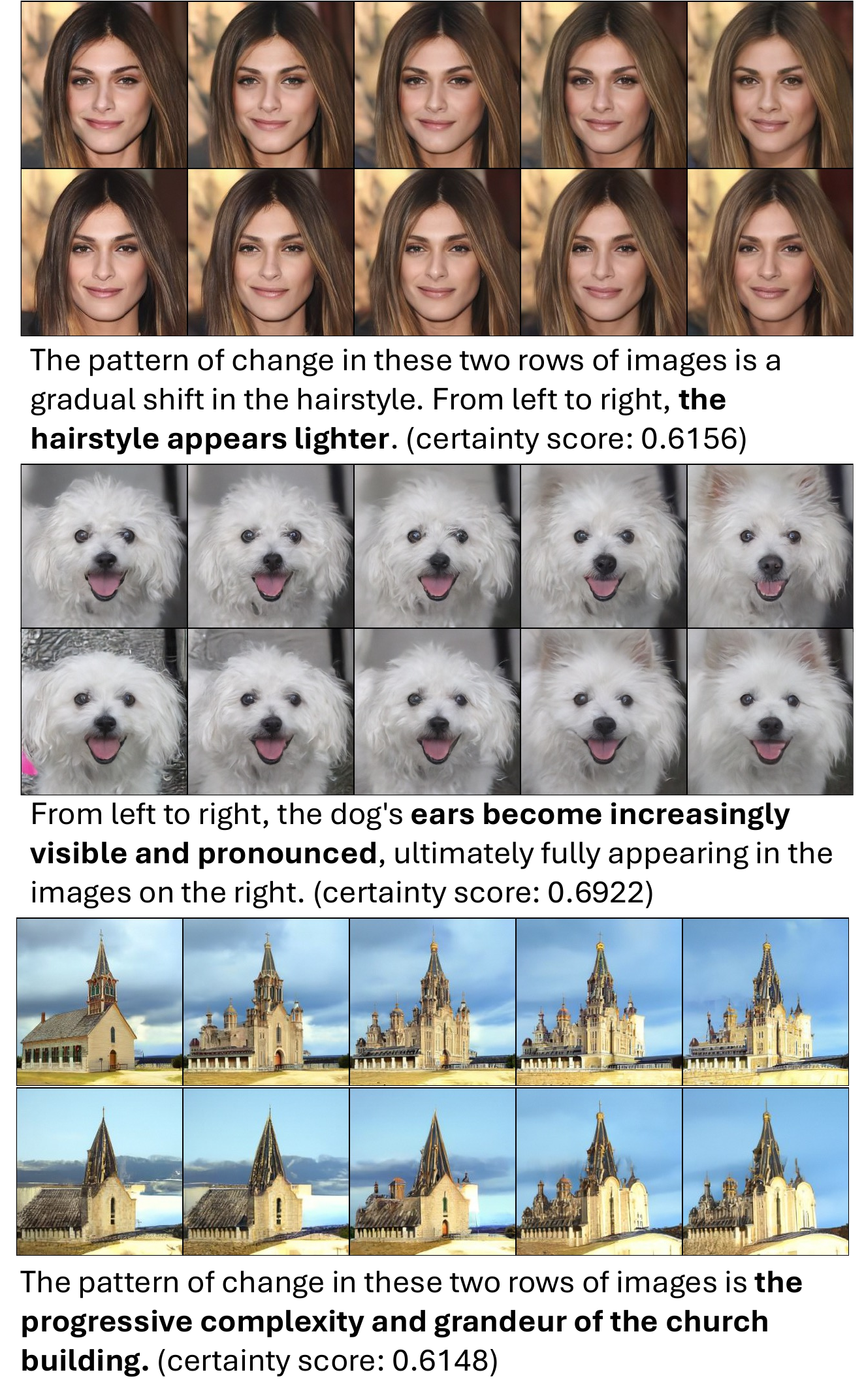}
\vspace{-5mm}
\caption{Example of generated explanations for DDPM under the disentanglement bias.}
\label{fig:example_disentangle}
\vspace{-5mm}
\end{center}
\end{figure}

\subsection{Qualitative Evaluation}
To analyze the explanations for DDPM under disentanglement bias, we manipulate the latent variable along a latent direction and compare it with the one that first traverses along another latent direction and then traverses along the same latent direction. The disentangled latent variable would be invariant with respect to the variations in another latent dimension. In Figure~\ref{fig:evaluation_disentanglement}, for each latent direction, we pass two image sequences and an inductive bias prompt based on the disentanglement formula to the MLLM to obtain a common latent explanation. In comparison, the explanations generated without the inductive bias prompt as shown in Figure~\ref{fig:evaluation_disentanglement_wo} tend to show "no clear explanation" or wrong explanations as they only align with one image sequence but do not reflect the common pattern in both image sequences in view of the inductive bias. The addition of the inductive bias prompts can assist with ruling out the variation effects of other latent variables to capture the actual meaning.  

We also qualitatively evaluate the explanations for CSVAE under combination bias. We transverse a latent variable and compare it with the one that first traverses another latent variable in another group and then traverses the same latent variable, which is similar to the disentanglement bias. We then compare with the one that first transverses another latent variable in the same group and then traverses the same latent variable. As Figure~\ref{fig:evaluation_combination} depicts, our model can clearly show the explanations of latent variable $z_1, z_2, z_3$ as the color of the ground, the background color, and the shape of the object. The effect of removing inductive bias prompts leads to no clear explanation like the disentanglement bias in Figure~\ref{fig:evaluation_combination_wo}.

In Figure~\ref{fig:evaluation_conditional}, we provide the "young appearance" prompt to Stable Diffusion under conditional bias, and the explanations of all three top latent directions reflect the meaning of youth. The addition of inductive bias prompts can better identify the relation with the property of interest to capture the actual meaning of latent variables. In comparison, the one without the inductive bias prompts in Figure~\ref{fig:evaluation_conditional_wo}, cannot find clear explanations or simply describe the characteristics in the image sequence, lacking an abstract generalization. More qualitative evaluation results can be found in Figure~\ref{fig:example_disentangle}.

\section{Conclusion}
In this paper, we introduced \textit{LatentExplainer}, a framework designed to generate semantically meaningful explanations of latent variables in deep generative models. Our work makes three key contributions: (1) Inferring the meaning of latent variables by translating inductive bias formulas into structured perturbations of latent variables through a coding agent; (2) Aligning explanations with inductive biases by converting mathematical formulations into textual prompts in MLLMs; (3) Introducing an uncertainty-aware approach that assesses explanation consistency. Quantitative and qualitative evaluations across multiple datasets and generative models demonstrate that \textit{LatentExplainer} significantly outperforms baseline methods. The incorporation of inductive bias prompts leads to more structured and meaningful explanations, while uncertainty-aware filtering further enhances consistency and reliability. Our findings highlight the importance of inductive bias prompting and uncertainty quantification in bridging the gap between generative models and human interpretability.

\section{GenAI Usage Disclosure}
We used generative AI tools to improve the grammar and clarity of the writing.

\begin{acks}
This work was supported by the National Science Foundation (NSF) Grant No. 2414115, No. 2403312, No. 2007716, No. 2007976, No. 1942594, No. 1907805, NIH R01AG089806, and NIH R01CA297856.
\end{acks}
%%
%% The next two lines define the bibliography style to be used, and
%% the bibliography file.
\newpage
\bibliographystyle{ACM-Reference-Format}
\bibliography{sample-base}

\end{document}